\pdfoutput=1

\documentclass[11pt]{article}
\usepackage[hyphens]{url}
\usepackage[final]{acl}

\usepackage{times}
\usepackage{latexsym}

\usepackage[T1]{fontenc}

\usepackage[utf8]{inputenc}

\usepackage{microtype}

\usepackage{inconsolata}

\usepackage{graphicx}
\usepackage{amsmath}
\usepackage{ascmac}
\usepackage{subfigure}

\title{A Judge-free LLM Open-ended Generation Benchmark Based on the Distributional Hypothesis}

\author{
 \textbf{Kentaro Imajo\textsuperscript{1,2,}$^*$},
 \textbf{Masanori Hirano\textsuperscript{1,2,}$^*$},
 \textbf{Shuji Suzuki\textsuperscript{1,2}},
 \textbf{Hiroaki Mikami\textsuperscript{1,2}}
\\
 \textsuperscript{1}Preferred Networks, Inc.,
 \textsuperscript{2}Preferred Elements, Inc.
\\
imos@preferred.jp, research@mhirano.jp, \{ssuzuki, mhiroaki\}@preferred.jp
}

\begin{document}
\maketitle
\def\thefootnote{*}\footnotetext{Equal contribution}\def\thefootnote{\arabic{footnote}}
\begin{abstract}
Evaluating the open-ended text generation of large language models (LLMs) is challenging because of the lack of a clear ground truth and the high cost of human or LLM-based assessments.
We propose a novel benchmark that evaluates LLMs using n-gram statistics and rules, without relying on human judgement or LLM-as-a-judge approaches.
Using 50 question and reference answer sets, we introduce three new metrics based on n-grams and rules: Fluency, Truthfulness, and Helpfulness.
Our benchmark strongly correlates with GPT-4o-based evaluations while requiring significantly fewer computational resources, demonstrating its effectiveness as a scalable alternative for assessing LLMs' open-ended generation capabilities.
\end{abstract}

\section{Introduction}

Large language models (LLMs) have recently demonstrated high capabilities across various tasks, particularly in open-ended text generation, as seen in ChatGPT \cite{chatgpt} and other models \cite{GPT4,touvron2023llama,Touvron2023,jiang2023mistral}.
In open-ended generation, LLMs must produce correct answers in a human-like style.
Thanks to scaling laws \cite{kaplan2020scaling,wei2022emergent,gunasekar2023textbooks}, this and many other tasks have been remarkably improved.

Evaluating LLMs' open-ended generation is crucial for their development but remains challenging.
The most reliable evaluation method is human judgment, such as in Chatbot Arena \cite{chiang2024chatbot}.
However, open-ended generation tasks lack a ground truth and clear objective criteria for evaluation.

Recent LLM-as-a-judge benchmarks\cite{MT-bench}, where a high-end LLM replaces human judges, partially address this issue but have limitations.
They require high-performing models, significant computational resources, and often produce non-deterministic results, varying with models and prompts.
Moreover, LLM-as-a-judge sometimes ignores mismatches in languages between questions and outputs, assigning high ratings despite the error.
Another limitation is that it cannot evaluate pre-training LLMs without instruction tuning, as these models often fail to produce human-like answers despite having open-ended generation capabilities.

Some argue that LLMs go beyond the distributional hypothesis\cite{harris1954distributional} owing to scaling laws, but we propose a different perspective.
While some studies claim the hypothesis is no longer valid or sufficiently explainable \cite{chiang2023distributional,enyan2024llms},
its applicability may be less apparent due to the high-dimensional and sparse nature of LLMs.
For example, early in the LLM era, MAUVE \cite{pillutla2021mauve} effectively measured how closely generated text aligned with human text distributions by analyzing the generated text embeddings.
This suggests the hypothesis remains relevant for evaluating generated text.

We propose a novel benchmark for assessing open-ended generation quality without human or LLM judgment, leveraging the distributional hypothesis.
Our approach assumes that the good generation of LLMs satisfied the good agreement of the word distribution with the desirable answer distribution for given questions (contexts).
Our benchmark then evaluates this alignment using deterministic n-gram-based metrics that are efficient and require minimal computational resources.

Results show that our benchmark reliably approximates LLM-as-a-judge evaluations with significantly lower computational costs.

All code and evaluation materials are publicly available on GitHub: \url{https://github.com/pfnet-research/pfgen-bench}.

\section{Related Work}
Recently, LLMs have advanced significantly.
High-performance models like ChatGPT \cite{chatgpt}, GPT-4 \cite{GPT4}, GPT-4o, and others have demonstrated notable improvements in performance and generalization.
These models are built on the transformer architecture \cite{Vaswani2017}, followed by models such as BERT \cite{Devlin2018} and the GPT series \cite{GPT-1, GPT-2, GPT-3}.
Additionally, various other models have emerged, including Bard \cite{bard}, Gemini, LLaMA \cite{touvron2023llama, Touvron2023}, Dolly \cite{dolly}, BLOOM \cite{scao2022bloom}, Vicuna \cite{vicuna}, and PaLM \cite{Chowdhery2022, Anil2023}.

Evaluating LLM performance is equally important.
For instance, lm\_eval \cite{eval-harness} is a benchmarking platform for assessing LLMs across multiple tasks.
Various benchmarks have been introduced to evaluate specific capabilities, such as MMLU \cite{hendryckstest2021} for general knowledge, MMLU-Pro \cite{wang2024mmlu} and GPQA \cite{rein2023gpqa} for reasoning, mathematical benchmarks \cite{cobbe2021gsm8k, hendrycksmath2021}, MGSM \cite{shi2022language} for multilingual reasoning, and HumanEval \cite{chen2021codex} and MBPP \cite{austin2021program} for programming proficiency.
Additionally, studies have examined GPT-4’s performance of \cite{GPT4} across domains such as accounting examinations \cite{Eulerich2023}, medicine \cite{Nori2023}, and law \cite{Iu2023, Choi2023}.

Beyond domain- or task-specific benchmarks, evaluations also focus on output quality, particularly in assessing AI assistants and question-answering interactions.
MT-bench \cite{MT-bench} introduced a benchmark for multi-turn conversations using LLM-as-a-judge, where a high-performance LLM evaluates responses.
The LLM-as-a-judge method is widely adopted.
Moreover, Chatbot Arena \cite{chiang2024chatbot} proposed a platform in which humans conduct continuous pairwise comparisons of model outputs.
Given the significance of these evaluations, this study introduces a new benchmark for assessing open-ended generation quality.

Although we use n-gram-based metrics, our approach is rooted on the distributional hypothesis \cite{harris1954distributional}.
While LLMs challenge this hypothesis, it has long supported language modelling.
Many embedding and language models have been developed based on this principle \cite{mikolov2013efficient,sarzynska2021detecting,bojanowski2017enriching,Devlin2018}, leading to the modern LLMs.
Despite arguments that the hypothesis cannot fully explain LLMs' generalization \cite{chiang2023distributional,enyan2024llms}, we assume it remains useful for evaluating LLM outputs.

\section{Proposed Benchmark}\label{sec:method}
Our benchmark was designed to measure the open-ended generation capability and common sense of LLMs, including pretrained models.
This benchmark also aims to eliminate the dependency on LLMs or prompts during evaluation, unlike LLM-as-a-judge approaches \cite{MT-bench}.
Because LLM-as-a-judge sometimes assigns unjustifiably high scores to answers in the wrong language---especially when the answer is not intended to be in English---we constructed this benchmark in Japanese.

The proposed benchmark was constructed in three major steps:
\begin{itemize}
    \setlength{\parskip}{0cm} 
    \setlength{\itemsep}{0cm} 
    \item The construction of questions and sample answers.
    \item The construction of reference answer sets,
    \item The design of evaluation score calculation for LLMs.
\end{itemize}
\begin{figure}[tbp]
    \centering
    \includegraphics[width=\linewidth]{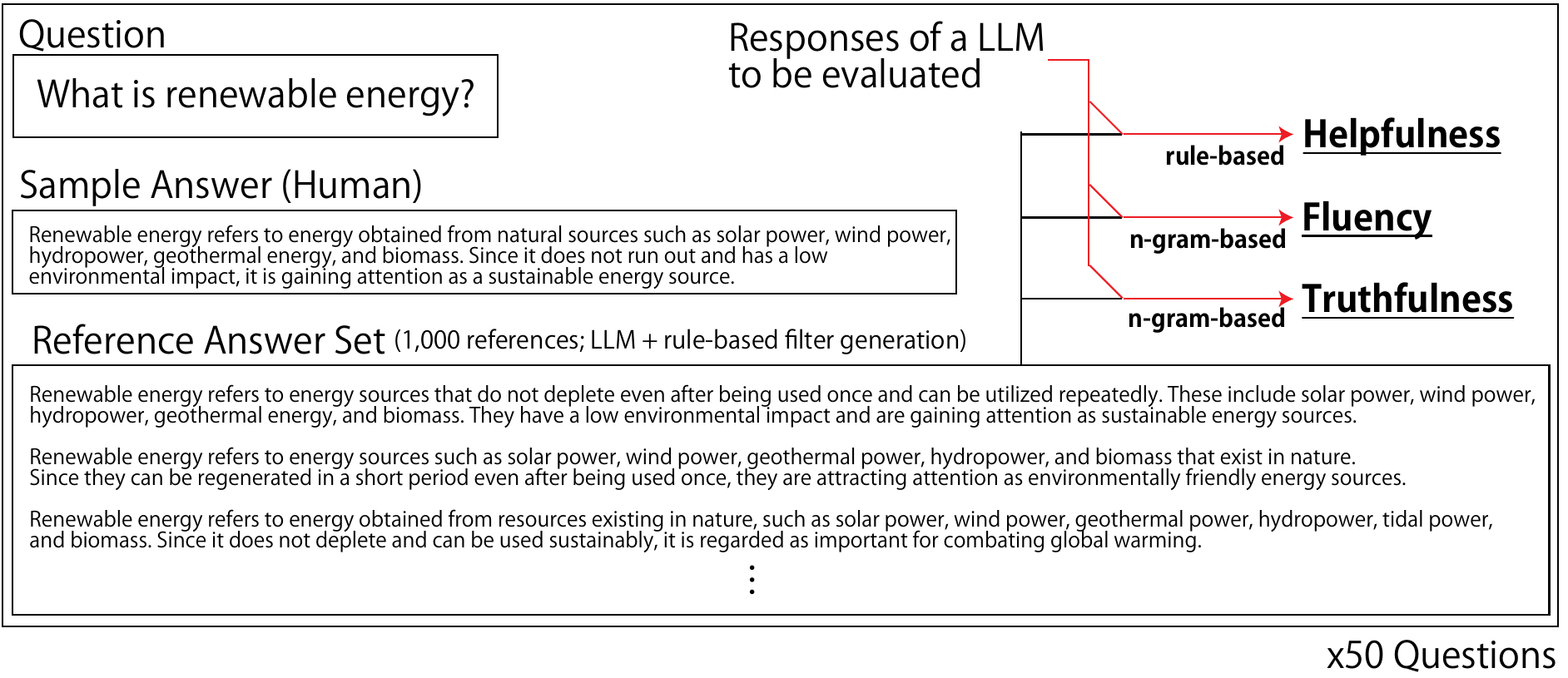}
    \vspace{-7mm}
    \caption{Evaluation outline}
    \label{fig:outline}
\end{figure}
Figure \ref{fig:outline} shows the evaluation flow in our benchmark.
For evaluating LLMs, the questions and reference answer sets were used.
Then, the evaluated LLMs generated outputs for each question, and these outputs along with the reference answer set were used to calculate scores.
Although this section explains all three steps, only the score calculation in the last step is necessary for evaluating LLMs.

\subsection{Questions and Sample Answers}
We referred to the national curriculum guidelines to construct 50 questions across multiple subjects that reflected common situations in the nation:
\begin{itemize}
    \setlength{\parskip}{0cm} 
    \setlength{\itemsep}{0cm} 
    \item Language: 4 questions
    \item Social Studies: 12 questions
    \item Mathematics: 4 questions
    \item Sciences: 16 questions
    \item Art and Culture: 8 questions
    \item Health: 4 questions
    \item Informatics: 2 questions
\end{itemize}
For example, in social studies, one question is "What is renewable energy?" 
All questions were designed to be concise and answerable in short form (approximately 100  Japanese characters, equivalent to 40 -- 50 words in English).
Additionally, we manually created sample answers.
For the above example, the manually created sample answer is "Renewable energy refers to energy obtained from natural sources such as solar power, wind power, hydropower, geothermal energy, and biomass. Since it does not run out and has a low environmental impact, it is gaining attention as a sustainable energy source."\footnote{Translated by GPT-4o}
Sample answers follow a style similar to that of official government documents, around 100 characters (roughly equivalent to 40 -- 50 words in English).

The number of questions reflected the volume of each subject’s subcategory.
These 50 questions were made available in an online repository, as mentioned earlier.

\subsection{Reference Answer Sets}
Because an open-ended generation question can have infinite possible answers, we constructed a reference answer set for each question consisting of many highly probable answers to define the answer distribution.

Ideally, these reference answers would be manually created by many people.
However, for practical reasons, we employ high-performance LLMs --specifically, Japanese-specific models.
For this purpose, we selected instructed LLMs with large parameter sizes (70B or 100B) available online (not via API) to ensure high performance and availability.
Although these LLMs require huge computational resources, note that this computational resource-intensive process is needed only for constructing the reference answer sets, not for evaluating each model, unlike LLM-as-a-judge.

Following these criteria and ensuring model family diversity, we adopted stockmark-100b\footnote{\url{https://huggingface.co/stockmark/stockmark-100b}}, PLaMo-100b\footnote{\url{https://huggingface.co/pfnet/plamo-100b}}, and Swallow-MX-8x7b-NVE-v0.1\footnote{\url{https://huggingface.co/tokyotech-llm/Swallow-MX-8x7b-NVE-v0.1}}, which were available at the time of construction.

The reference answer set was constructed using the following steps:
\begin{enumerate}
    \setlength{\parskip}{0cm} 
    \setlength{\itemsep}{0cm} 
    \item Generate 1 million responses per question using many-shot prompting for each of the three models
    \item Remove responses exhibiting hallucinations using rule-based filtering
    \item Remove responses exhibiting hallucination using frequency filtering
    \item Refine the responses to 1000 per question based on length and representativeness
\end{enumerate}
These steps were designed to gather representative, accurate answers while excluding erroneous or unexpected outputs.
Detailed explanations of these steps follow.

\subsubsection{Step 1: 1 Million Responses Generation}\label{sec:generate-answer}
To generate responses similar to the sample answers in writing style and length, we employed many-shot Q\&A generation.
The generation prompt provided explicit instructions for both style and length, and 20-shot question-and-answer examples drawn from the other 49 questions.

The actual prompts are shown in Appendix \ref{appendix:generation-prompt}.

We used the following generation parameters to encourage varied responses: temperature = 1.0, top\_p = 0.98, and top\_k = 1000.

For the three questions that the selected LLMs could not answer correctly, we incorporated a retrieval-augmented generation.
We extracted 50 relevant sentences from web pages via Google searches and randomly selected five sentences per prompt to ensure diverse responses.

Based on these settings, we generated 1.5 billion responses across stockmark-100b, PLaMo-100b, and Swallow-MX-8x7b-NVE-v0.1 for 50 questions.
To align responses with the official writing style, we employed text normalization using regular expressions following the guidelines\footnote{\url{https://www.bunka.go.jp/seisaku/bunkashingikai/kokugo/hokoku/pdf/93651301_01.pdf}}.

\subsubsection{Step 2: Rule-based Filtering}
Rule-based filters were designed for each question to eliminate responses containing hallucinations.
For example, for the question "How many times do the hour and minute hands of a clock overlap in a day?," the correct answer should include "22".
If a response contained "11" or "23, " it was removed.
To implement this, responses for each question were manually reviewed, and erroneous responses were filtered based on predefined rules.

\subsubsection{Step 3: Frequency-based Filtering}\label{sec:del-hal}
To further refine the response sets, we also employed a 5-gram frequency-based filter.
While the rule-based filtering in the previous section eliminates responses owing to LLM misunderstanding or mislearning, LLMs sometimes generate sentences that humans typically would not write. (akin to the Type I error defined in \cite{pillutla2021mauve}.)

This filter removes responses containing any 5-gram that appears only once among all responses to a given question.
For example, for the question "Tell me about the easternmost, westernmost, northernmost, and southernmost points of Japan," the filter would remove an incorrect response like "The easternmost point of Japan is Minamitorishima (153\textdegree 59'30"E, 20\textdegree 43'57"N)" (the correct coordinates being 153\textdegree 59'12"E, 24\textdegree16'59"N).
Because many errors were not be captured by the rule-based filters, the 5-gram frequency filter effectively removes hallucinations.

\subsubsection{Step 4: Response Refining}
In the final step, we refined the generated reference answers to 1,000 responses per question.
First, we reduced the set to 30,000 responses per question by sorting them based on their proximity to 100 characters in length, resulting in a reference answer set with an average of 100 characters (standard deviation: 3.4 characters).

Next, we further refined the set to 1,000 responses per question based on word distribution diversity.
We employed a 1--10-gram distribution and used a hill-climbing optimization to minimize the mean squared error between the 1--10-gram distribution of the final 1,000 responses and that of the original 30,000.

\subsection{Evaluation Score Calculation}
Using the reference answer sets from the previous section, we computed benchmark scores for the target LLMs.

We generated responses from the target LLMs using the same prompt as in Section \ref{sec:generate-answer} (see Appendix \ref{appendix:generation-prompt}).
The temperature was not fixed (varied by LLM), and the number of responses generated was flexible.
Increasing the temperature increased output randomness, whereas generating more responses improved convergence of the evaluation scores.
Thus, any setting is acceptable provided that repeated trials yield low variance.
Empirically, the evaluation scores converged well unless the temperature was excessively high.
In our experiments, we used a temperature of 1.0 and generated 100 responses for on-premise models (e.g., those on Hugging Face).
For API accessed models, such as OpenAI’s, we set the temperature to 0.0, and limited generations to three for cost reasons.

Next, we computed three evaluation metrics using the reference answer sets and  generated responses:
\begin{itemize}
    \setlength{\parskip}{0cm} 
    \setlength{\itemsep}{0cm} 
    \item Fluency: The inner product of the occurrence ratios of 10-grams.
    \item Truthfulness: The proportion of 3-grams with an occurrence frequency of at least 0.5\%.
    \item Helpfulness: Evaluation based on manually defined rules.
\end{itemize}
Fluency scores were normalized so that the evaluation score of the reference answer set was 1.0.
The final benchmark score is the average of these three metrics.
The following sections describe each calculation in detail.

\subsubsection{Fluency}
Fluency is defined here as a metric that checks whether an LLM response is written fluently.

Fluency was assessed by converting the responses into character-level n-grams (1--10 grams) and checking their likelihood in the reference answer set.
We extracted all character-level n-grams from the generated responses, computed their occurrence frequencies in the reference answer set, and summed them.

We note that we employed character-level analysis because this benchmark currently targets Japanese.
Unlike English, which is alphabet-based, Japanese has ideographic characteristics; hence, the character-level analysis is preferable over word-level analysis (for English, the word or token level may be preferable).

Because responses are expected to be approximately 100 characters long, longer responses tend to have higher scores because of the summing of occurrence frequencies.
To prevent this, we introduce a linear discounting mechanism reduces the score beyond 100 characters and reaches zero at 150 characters.

In other words, let a response consist of $L$ characters denoted by $C_1, C_2, \dots, C_L$.
We explicitly add beginning-of-sentence and end-of-sentence tokens, resulting in $C_0, C_1, \dots, C_L, C_{L+1}$.
A character-level $w$-gram is defined as $G^{w}_i = \{C_i, C_{i+1}, \cdots, C_{i+w-1}\}$ and a response constructs $L-w+3$ character-level $w$-grams.
Let the likelihood of the occurrence of $G^{w}_i$ in the reference answer set be ${L}^{w}_{G^{w}_i} \in [0, 1]$.
The Fluency score before discounting is defined as
\begin{equation}
    F^\star_w := \sum_{i=0}^{L-w+2} {L}^{w}_{G^{w}_i}, 
\end{equation}
which corresponds to the inner product of the frequency vector of the response’s character-level $w$-grams and the probability vector of the reference answer set’s $w$-grams.

To prevent the score from increasing monotonically with $L$, we apply a discount factor
\begin{equation}
    F_w := \left(1 - \frac{\max(L-100, 0)}{50}\right)\sum_{i=0}^{L-w+2} {L}^{w}_{G^{w}_i}
\end{equation}
where $F_w = 0$ when $L \geq 150$ and $F_w = F^\star_w$ when $L \leq 100$.

The final fluency score is
\begin{equation}
    F := \sum_{w=1}^{10} F_w.
\end{equation}
This design ensures that responses of approximately 100 characters receive the highest scores.

Assuming that the occurrence likelihood ${L}^{w}_{G^{w}_i}$ of $w$-gram is, on average, constant, we approximate
\begin{align}
    &\sum_{i=0}^{L-w+2} {L}^{w}_{G^{w}_i} \sim L-w+3,\\
    &F_w \sim \left(1 - \frac{\max(L-100, 0)}{50}\right)(L-w+3).
\end{align}
Therefore, $F_w$ exhibits a peak at $L=100$.
Since this approximation applies to all $w$ values, the fluency scores exhibit the same behavior.

\subsubsection{Truthfulness}
Truthfulness is defined as a metric that measures how accurately a response provides correct information.
Similar to Section \ref{sec:del-hal}, we assume that the appearance of an extremely low-frequency character-level n-gram is likely due to hallucinations or incorrect outputs.

For this metric, we define character-level 3-grams with an occurrence frequency below 0.5\% in the reference answer set as unlikely, and we compute the proportion of 3-grams with a frequency of at least 0.5\%.
Because this index indicates the presence of hallucinations, punctuation marks and brackets are excluded from the frequency calculation.

Similar to Fluency, we introduce a discount for responses exceeding 100 characters; however, for Truthfulness, we allow truncation at the optimal length that maximizes the score.
In other words, if a response exceeds 100 characters, the score is determined by truncating it to the length that yields the highest Truthfulness score.
This approach helps absorb randomness in outputs from smaller models that may struggle to generate text of an appropriate length.

We formally define the character-based 3-gram as $G^{3}_i = \{C_i, C_{i+1}, C_{i+2}\}$.
Let the occurrence likelihood of the 3-grams in the reference answer set be ${L}^{3}_{G^{3}_i} \in [0, 1]$, and let $\Omega_I = \{i | C_i \notin \{punctuation, symbols\}\} \subset \{1, 2, \cdots, I\}$ ($100 \leq I \leq L$ or $I=L$) denote the set of indices corresponding to non-punctuation characters.

The Truthfulness score before discounting is defined as
\begin{align}
    T^\star_{I} &:= \frac{1}{|\Omega_{I}|}\sum_{i \in \Omega_{I}}\frac{\min\left({{L}^{3}_{i}}^\star, 0.005\right)}{0.0005},\\
    {{L}^{3}_{i}}^\star &:= \max\left({L}^{3}_{G^{3}_{i-1}}, {L}^{3}_{G^{3}_{i}}, {L}^{3}_{G^{3}_{i+1}}\right).
\end{align}
Please note that technically, this is not exactly the same but almost the same as the proportion of 3 grams with a frequency of at least 0.5\%.

Because $T^\star_{I}$ increases with $I$ owing to the growing size of $|\Omega_{I}|$ (similar to the Fluency score $F^\star$), a discount is applied.
The temporal Truthfulness score corresponding to output length $I$ is defined as
\begin{equation}
    T_{I} := \left(1 - \frac{\max(I-100, 0)}{50}\right)T^{\star}_{I}.
\end{equation}

Finally, because the final Truthfulness Score is defined as the maximum temporal score obtained by truncating the response to $I$ characters, it is computed as
\begin{equation}
    T := \left\{\begin{array}{ll}
       \left(1 - \frac{\max(L-100, 0)}{50}\right)T^{\star}_{L} ~~~~~~~~~~(L\leq 100) \\
       \underset{I \in \{100, 101, \cdots, L\}}{\max} \left(1 - \frac{\max(I-100, 0)}{50}\right)T^{\star}_{I}\\  ~~~~~~~~~~~~~~~~~~~~~~~~~~~~~~~~~~~~~~~~~~~~~~~~(100 \leq L)
    \end{array}\right..
\end{equation}
As $|\Omega_{I}|$ is approximately equal to $I$, $T_{I}$ is maximized when $I=100$, as with Fluency.
Thus, Truthfulness effectively reflects the ratio of 3 grams that occur at a certain frequency in the reference answer set, serving as a metric for the presence of hallucinations in the first 100 characters.

\subsubsection{Helpfulness}
Helpfulness is designed to measure how well a response includes the necessary information based on predefined and manually defined rules.
For each question, we identified a set of key terms that should appear in a correct response and calculated the proportion of these terms present.
For example, for the question "What is superconductivity?," essential terms such as "temperature," "resistance," "zero/0," and "magnetic" are required.
In this example, all rules for helpfulness are defined by AND/OR conditions.
These important keywords were treated equally in most cases, a small number of questions assigned different weights to certain terms based on their relative importance (see more details in our GitHub repository).

For responses exceeding 100 characters, we compute the score by truncating the response at various lengths $I \geq 100$ and applying a discount factor $\left(1 - \frac{\max(I-100, 0)}{50}\right)$.
The highest score obtained after truncation is selected as the final Helpfulness score.
This approach ensures that longer responses, which may contain additional relevant terms, are not unfairly penalized.

\section{Experiments}
While Section \ref{sec:method} outlines the details of the proposed benchmark, this section evaluates its effectiveness by addressing two key research questions that were not explicitly clarified earlier.

First, it remains uncertain whether the proposed benchmark methodology effectively measures LLM performance.
Although the previous section introduced three evaluation metrics, their validity, whether mathematically or theoretically, was not demonstrated.
To address this, we examine the correlation between our benchmark scores and existing evaluation methodologies, particularly LLM-as-a-judge \cite{MT-bench}.

Second, the balance and comprehensiveness of the 50 benchmark questions need to be assessed. It is unclear whether this question set provides a well-rounded evaluation of LLM capabilities.
To investigate this, we analyze correlations between our benchmark and established benchmarks, ensuring that our covers key aspects of LLM performance in alignment with widely accepted benchmarks.

The following sections present our preliminary analyses and experiments to answer these two questions.

\subsection{Ex. 1: Benchmark Characterization}
We perform a preliminary analysis to examine the benchmark’s characteristics.
This experiment visualizes the score distribution for each evaluation metric (Fluency, Truthfulness, and Helpfulness) across multiple LLMs to explore metric relationships.

Additionally, we assess the stability of reference answer sets.
As outlined in Section \ref{sec:method}, these sets were constructed using three high-performance Japanese-specific LLMs.
To evaluate stability, we compare benchmark scores computed with an ensemble of all three models against those using a single model.
If scores vary significantly by reference model, it may indicate instability in the construction method.
These analyses clarify our benchmark’s fundamental properties robustness.

\subsection{Ex. 2: Comparison with LLM-as-a-Judge}
We compared our benchmark results with those of LLM-as-a-judge \cite{MT-bench} for multiple recognized LLMs.
OpenAI's GPT-4o (2024-05-13 ver.) evaluated responses for 50 questions on a 10-point scale.
For LLMs supporting both chat completion and completion, we tested both.

We then measured the correlation between our benchmark scores and LLM-as-a-judge scores.

A detailed prompt is provided in Appendix \ref{appendix:llm-as-a-judge-prompt}.

\subsection{Ex. 3: Comparison with Existing Bench.}
In the final experiments, we tested whether the proposed benchmark aligned with existing benchmarks.

We used two: the Japanese MT-Bench \footnote{\url{https://github.com/Stability-AI/FastChat/tree/jp-stable/fastchat/llm_judge}} and Nejumi LLM Leaderboard 3\footnote{\url{https://api.wandb.ai/links/wandb-japan/yd28xzvv}}, both widely used for Japanese LLM evaluation.

For a fair comparison, we assessed only the 37 models appearing in both our benchmark and the two existing ones.
If an LLM supported both chat completion and completion, we used the higher score for correlation calculations.

\section{Results}
\subsection{Ex. 1: Benchmark Characterization}
\begin{figure*}[htbp]
    \centering
    \subfigure[Fluency and Truthfulness]{\includegraphics[width=0.32\linewidth]{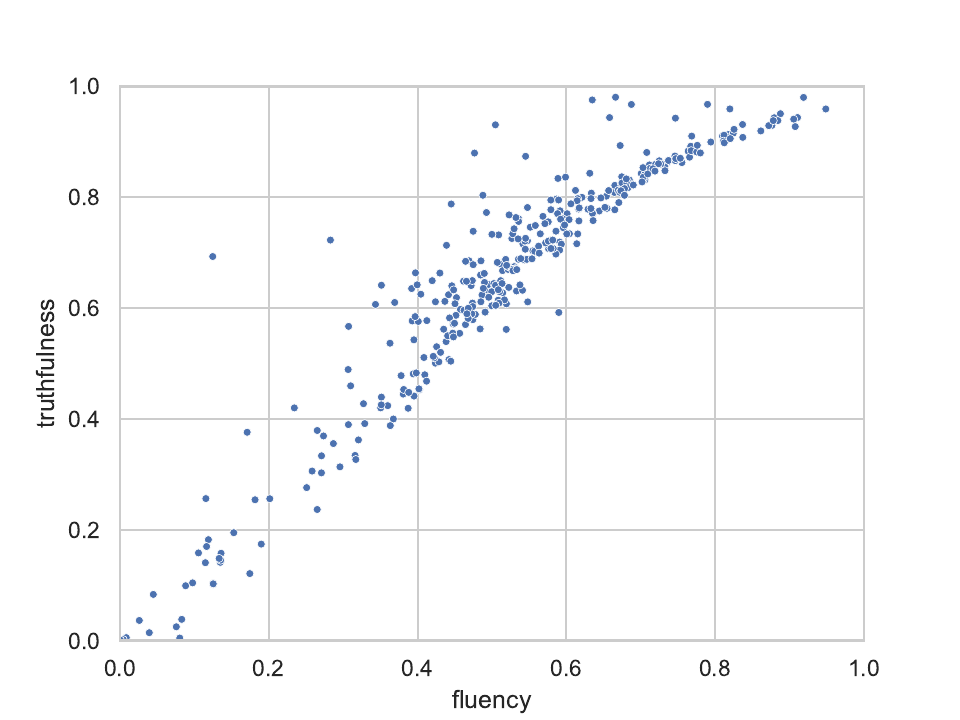}}
    \subfigure[Fluency and Helpfulness]{\includegraphics[width=0.32\linewidth]{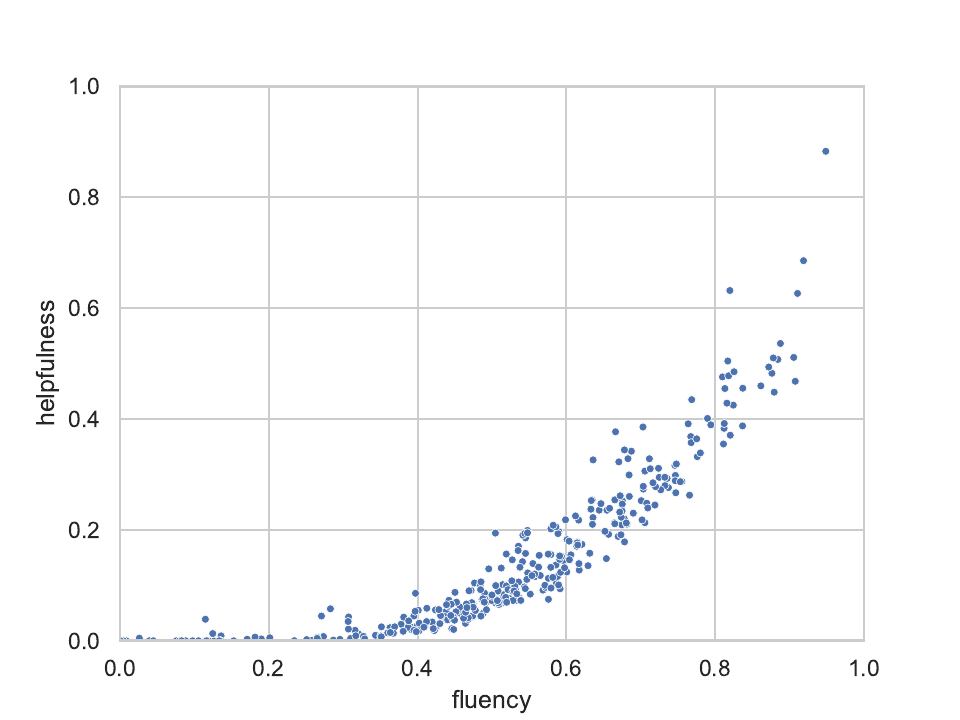}}
    \subfigure[Truthfulness and Helpfulness]{\includegraphics[width=0.32\linewidth]{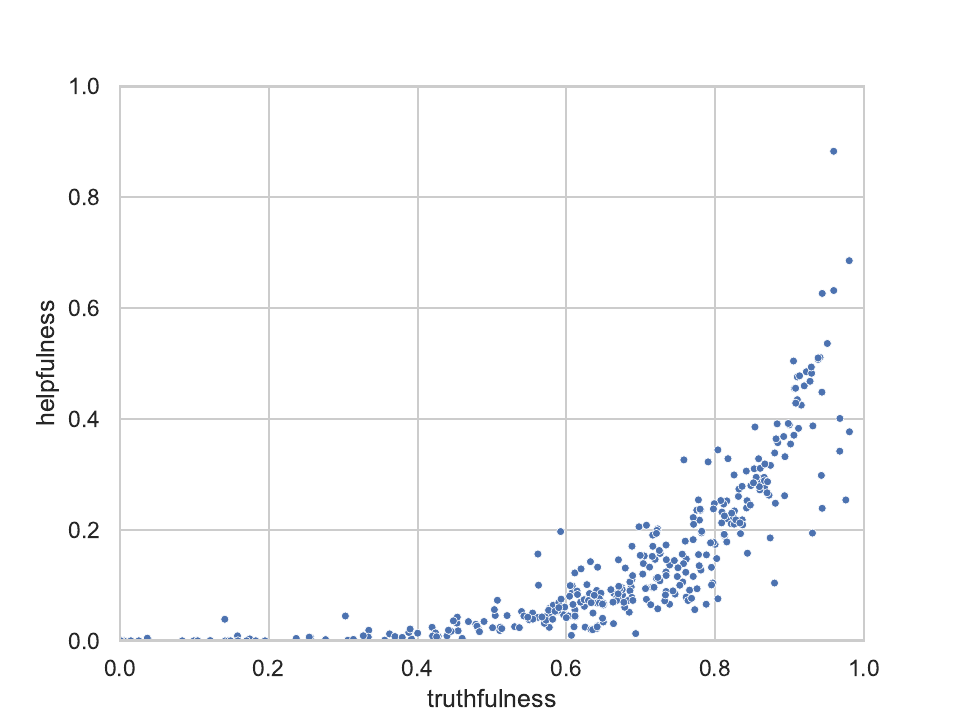}}
    \vspace{-3mm}
    \caption{Relationships between Fluency, Truthfulness, and Helpfulness}\label{fig:fluency_truthfulness_helpfulness}
\end{figure*}

Figure \ref{fig:fluency_truthfulness_helpfulness} shows relationships among Fluency, Truthfulness, and Helpfulness.
A list of benchmark scores for major LLMs is in Appendix \ref{appendix:score-table}.

The ranking of LLMs' remained generally consistent across metrics, although  some fluctuations occurred.
For Fluency and Truthfulness, certain LLMs already achieved near-perfect scores, suggesting they are close to their performance ceiling, while many still fall short for Helpfulness.

Next, we analyzed a reference answer set construction.
We used three models available at the time: stockmark-100, PLaMo-100b, and Swallow-MX-8x7b-NVE-v0.1, employing an ensemble approach.
We examined score variations to determine whether using a single model, rather than an ensemble, significantly affected benchmark results.

\begin{figure*}[htbp]
    \centering
    \subfigure[Plamo vs Swallow]{\includegraphics[width=0.32\linewidth]{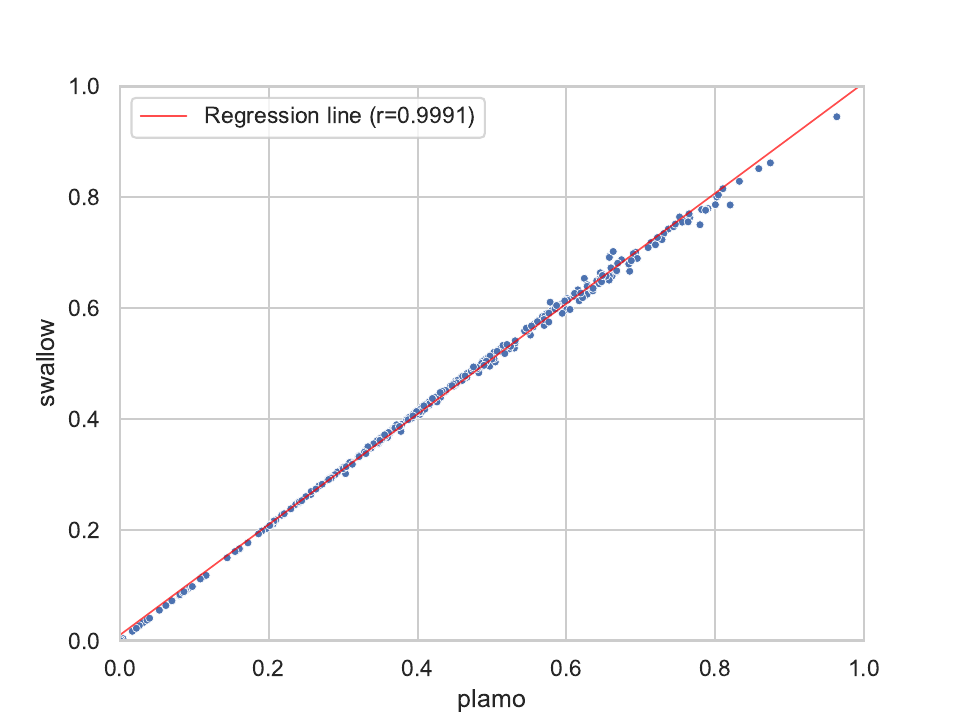}}
    \subfigure[Plamo vs Stockmark]{\includegraphics[width=0.32\linewidth]{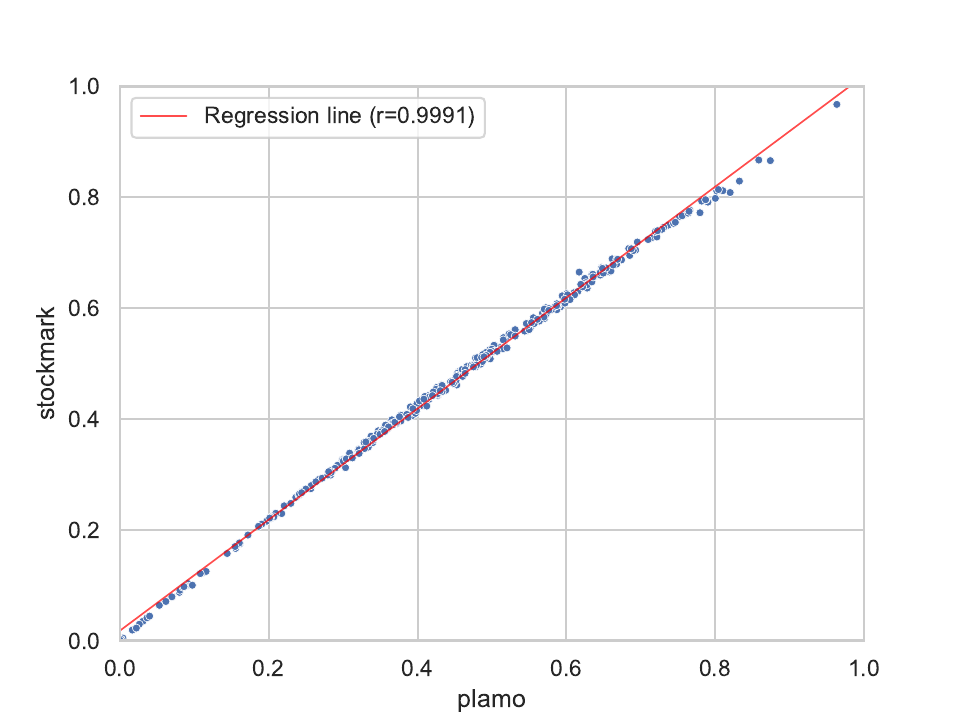}}
    \subfigure[Swallow vs Stockmark]{\includegraphics[width=0.32\linewidth]{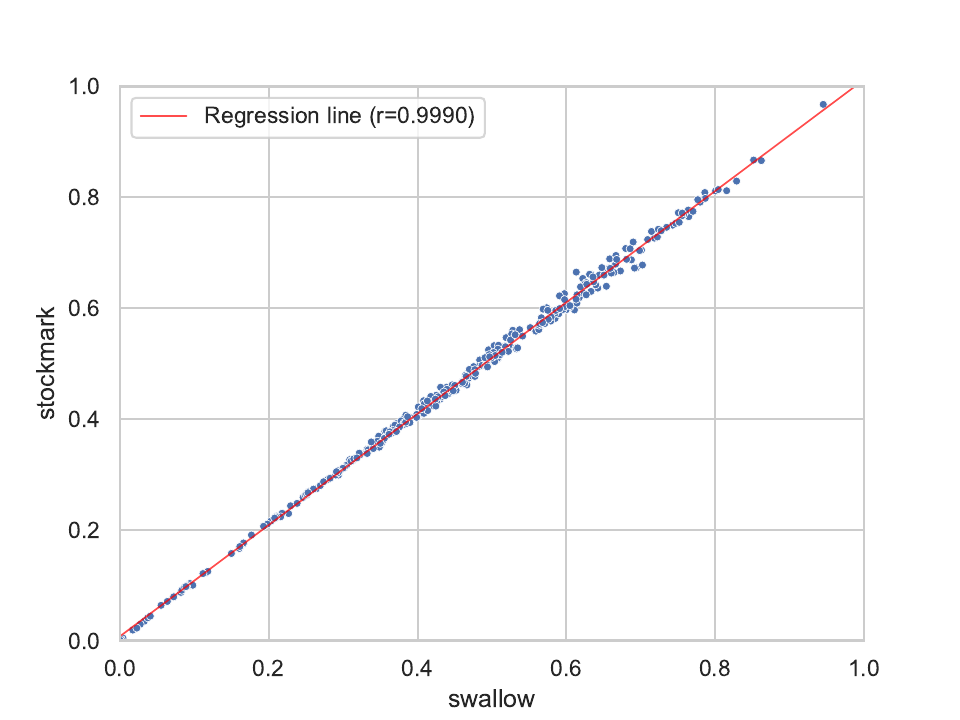}}
    \vspace{-3mm}
    \caption{Benchmark score comparison between using one of three LLMs to construct a reference answer set}\label{fig:plamo_swallow_stockmark}
\end{figure*}

Figure \ref{fig:plamo_swallow_stockmark} confirms a correlation above 0.999, indicating performance variation across models used to construct the reference set is small.
This suggests the reference set is stable and not highly dependent on any single model.

\subsection{Ex. 2: Comparison with LLM-as-a-Judge}

\begin{figure}[tbp]
    \centering
    \includegraphics[width=0.8\linewidth]{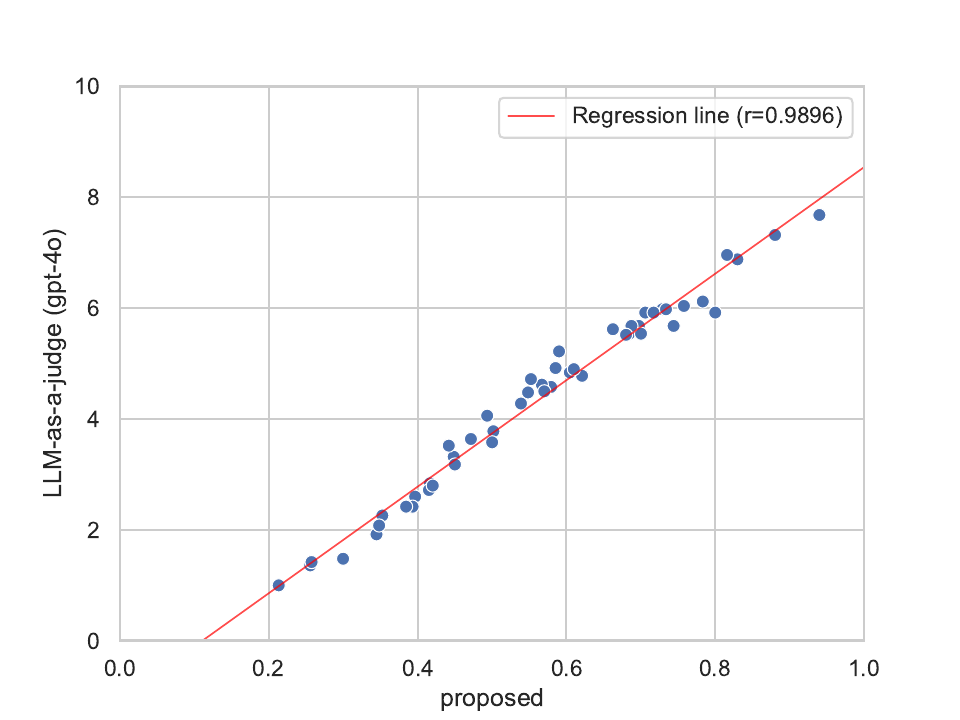}
    \caption{Comparison between our benchmark and LLM-as-a-judge}
    \label{fig:enter-label}
\end{figure}

Figure \ref{fig:enter-label}  presents the comparison between our benchmark and GPT-4o’s LLM-as-a-judge.
Owing to API costs, we limited the number of evaluated LLMs to 50.

The results show a high correlation of 0.9896, indicating that our benchmark performs comparably to LLM-as-a-judge.
While scores above 1.0 (above 9 by GPT-4o) and below 0.2 (near 0 by GPT-4o) were not observed, our benchmark’s resolution appears sufficient for practical use.

\subsection{Ex. 3: Comparison with Existing Bench.}

\begin{figure}[tbp]
    \centering
    \includegraphics[width=0.8\linewidth]{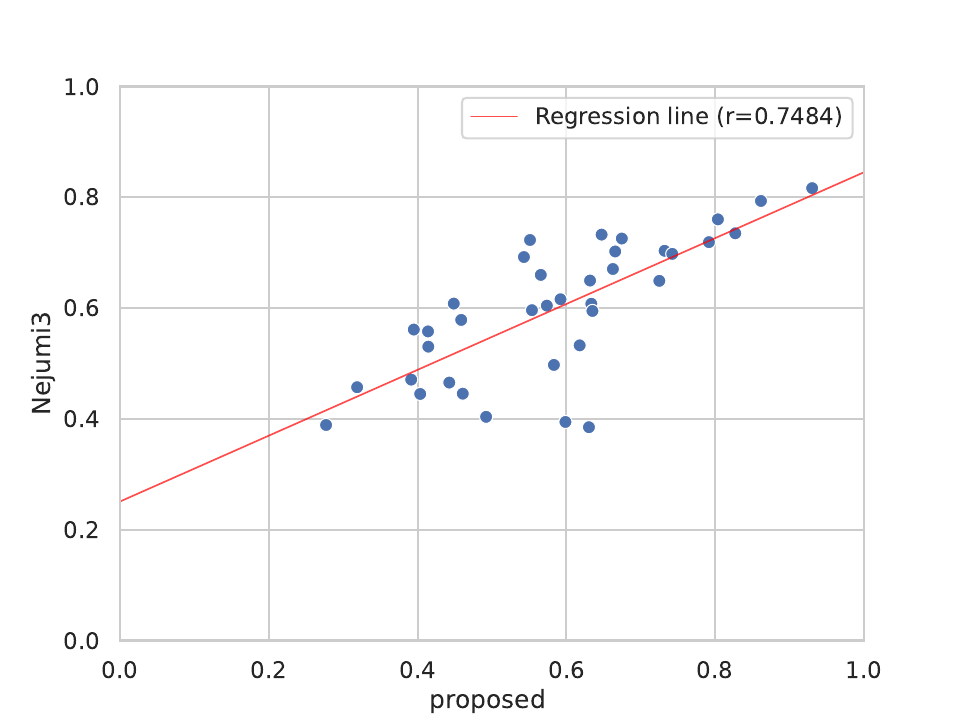}
    \caption{Comparison between our benchmark and Nejumi LLM Leaderboard 3}
    \label{fig:pfgen-nejumi3}
\end{figure}
\begin{figure}[tbp]
    \centering
    \includegraphics[width=0.8\linewidth]{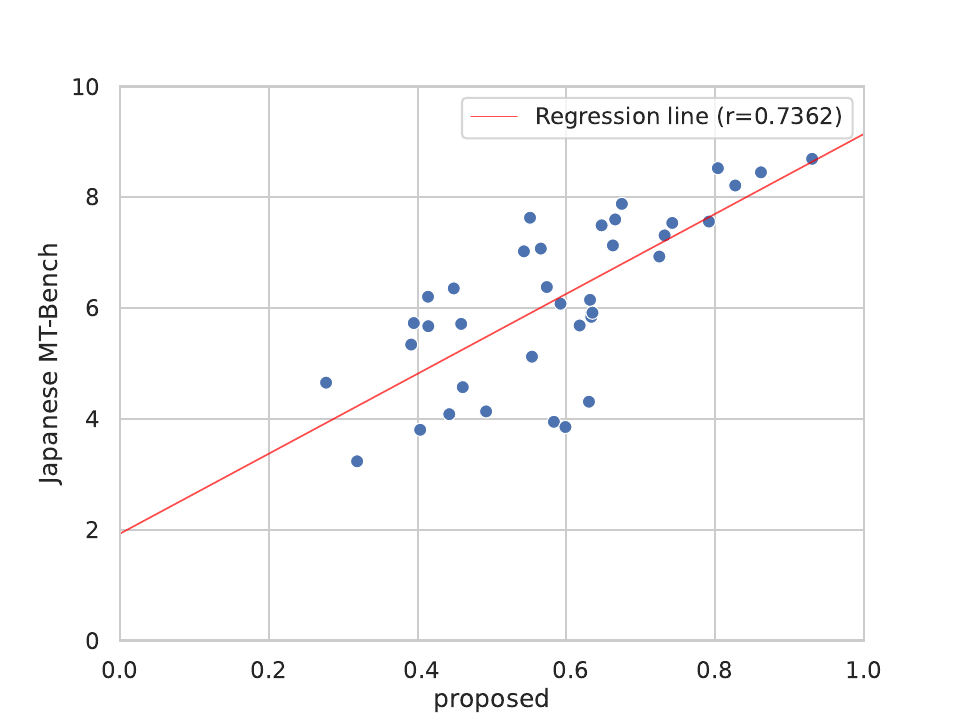}
    \caption{Comparison between our benchmark and Japanese MT-Bench}
    \label{fig:pfgen-mtbench}
\end{figure}

Figures \ref{fig:pfgen-nejumi3} and \ref{fig:pfgen-mtbench} present score comparisons between our benchmark and Nejumi LLM Leaderboard 3 or Japanese MT-bench.

The results confirm relatively high correlations above 0.7 for both benchmarks.
However, compared to the LLM-as-a-judge analysis, the correlation was slightly lower, likely because of task differences.
However, for high-performing LLMs, benchmark scores remained consistent across all benchmarks, suggesting our benchmark is robust and effective.

\section{Discussion}
The experimental results suggest that the proposed benchmark effectively captures the open-ended generation performance of LLMs.
In particular, our benchmark showed a high correlation with LLM-as-a-judge using GPT-4o.
Because LLM-as-a-judge requires significant computational resources for model evaluation, our benchmark, which does not require large calculations during evaluation, offers a beneficial and cost-effective alternative.

Surprisingly, some LLMs (e.g., anthropic/claude-3-5-sonnet-20240620 and openai/gpt-4o) achieved higher scores than the reference answer set and the three models used to construct the reference answer set (see Appendix \ref{appendix:score-table}). 
This suggests that the reference answer set likely captures a manifold of correct or desirable responses, rather than a single "gold answer."
Moreover, our n-gram-based scoring method was shown to effectively capture the alignment with this manifold.

Our results also support the validity of the distributional hypothesis, at least for LLM evaluation.
The n-gram-based distributional analysis was surprisingly robust for evaluating open-ended generations.
Several factors likely explain why character-level n-gram statistics are sufficient for this benchmark.
Owing to the short answer requirement (approximately 100 characters in our work) and the specific question types, the n-gram space of desirable responses is limited.
Although LLMs may eventually exceed the distributional hypothesis, the restricted statistics of the responses could still align with the hypothesis.
Because the n-gram-based analysis has limitations in its representation, future high-performing LLMs may not be adequately evaluated by our benchmark.
However, because our benchmark can evaluate models beyond those used in the reference set construction, the current evaluation would likely cover moderate performance improvements.

Our benchmark suggests a potential paradigm shift in LLM evaluations, but also highlights current limitations.
The success of our n-gram-based approach indicates that it could be expanded to general open-ended generation evaluations.
Currently, we have only tested strict conditional, open-ended generations.
This means our evaluation method is limited to Q\&A tasks, which have constrained answer spaces.
However, many tasks do not have concrete answers, such as idea generation or muti-turn human-like conversations.
Thus, we should explore how n-gram-based evaluations can be adapted to accommodate more complex tasks.
In addition to practical future work, we also need further mathematical or theoretical analyses connecting our benchmark to the distributional hypothesis.

\section{Conclusion}
In this study, we proposed a new benchmark to evaluate the open-ended generation performance of LLMs.
The benchmark consisted of 50 questions, with a sample answer and a reference answer set constructed for each question.
The reference answer set represents the desirable response distribution and is generated using multiple LLMs along with rule- and n-gram-based filters.
Moreover, the metrics, Fluency, Truthfulness, and Helpfulness were newly defined to evaluate LLMs.
Fluency and Truthfulness are based on n-gram calculations, while Helpfulness consists of manually created rules.
By combining these metrics, we developed a new benchmark for assessing LLMs.
In our experiments, we first confirmed that our proposed metrics have a high correlation with LLM-as-a-judge evaluated by GPT-4o.
A strong correlation with existing benchmarks was also observed.
These results indicate that n-gram-based evaluation for open-ended generation in LLMs is effective, likely because of the distributional hypothesis. However, further investigation is needed to explore this connection in future work.


\section*{Acknowledgments}
We thank Toshiki Kataoka for assisting in the review of sample answers to ensure  their accuracy during dataset preparation.

\bibliography{custom}

\appendix

\section{Generation Prompt for Reference Answer Sets}\label{appendix:generation-prompt}
All sentences in the examples are translated by GPT-4o.
\begin{itembox}[l]{\textbf{Example for chat Completion}}
\{"role": "system", "content": "Answer the user's question in one line, in the same style and length as the example.\textbackslash n\textbackslash n\#\#  Example Answer\textbackslash nQ: What is renewable energy?\textbackslash nA: Renewable energy refers to energy obtained from natural sources such as solar power, wind power, hydropower, geothermal energy, and biomass. Since it does not run out and has a low environmental impact, it is gaining attention as a sustainable energy source.\textbackslash n..."\}, \{"role": "user", "content": "Q: What is Montesquieu's separation of powers system?"\}
\end{itembox}
\begin{itembox}[l]{\textbf{Example for Completion}}
Answer the user's question in one line, in the same style and length as the example.\\\\
\#\# Example Answer\\
Q: What is renewable energy?\\
A: Renewable energy refers to energy obtained from natural sources such as solar power, wind power, hydropower, geothermal energy, and biomass. Since it does not run out and has a low environmental impact, it is gaining attention as a sustainable energy source.\\
...\\
Q: What is Montesquieu's separation of powers system?\\
A:
\end{itembox}

\section{LLM-as-a-Judge Prompt for Ex. 2}\label{appendix:llm-as-a-judge-prompt}
\begin{itembox}[l]{\textbf{LLM-as-a-Judge Prompt}}
\lbrack Instruction\rbrack\\
Please act as an impartial judge and evaluate the quality of the response provided by an AI assistant to the user question displayed below. Your evaluation should consider correctness and helpfulness. You will be given a reference answer and the assistant's answer. Begin your evaluation by comparing the assistant's answer with the reference answer. Identify and correct any mistakes. Be as objective as possible. The expected language is Japanese. Responses in languages other than Japanese will incur score deductions unless specifically required. Failure to use Japanese at all will result in the lowest evaluation. However, using Japanese is not mandatory when providing only Python scripts or calculation results, where Japanese is not essential. Additionally, your explanation of judgement should be in Japanese. After providing your explanation, you must rate the response on a scale of 1 to 10 by strictly following this format: "[[rating]]", for example: "Rating: [[5]]".\\
\\
\lbrack Question\rbrack\\
\{\{Question\}\}\\
\\
\lbrack The Start of Reference Answer\rbrack\\
\{\{Sample Answer\}\}\\
\lbrack The End of Reference Answer\rbrack\\
\\
\lbrack The Start of Assistant's Answer\rbrack\\
\{\{Model Response\}\}\\
\lbrack The End of Assistant's Answer\rbrack
\end{itembox}

\section{The Benchmark Scores for Some Major LLMs}\label{appendix:score-table}
See table \ref{tab:score-table}.
\begin{table*}[htbp]
    \centering
    \caption{List of scores for major models. All results can be found at the URL \url{https://github.com/pfnet-research/pfgen-bench}.}
    \begin{tabular}{cc|cccc}
    model & type & Score & Fluency &	Truthfulness &	Helpfulness\\\hline
    Sample Answer (Human) & N/A & 1.0501 & 1.155 & 0.996 & 1.000 \\
    anthropic/claude-3-5-sonnet-20240620 & chat & 0.9303 & 0.949 & 0.959 & 0.883\\
    openai/gpt-4o & chat &  0.8615 & 0.919 &	0.980 &	0.686\\
    Reference Answer Set & N/A & 0.8494 & 0.936 & 0.978 & 0.505\\
    openai/gpt-4 & chat & 0.7916 & 0.888 & 0.951 & 0.536\\
    tokyotech-llm/Swallow-70b-NVE-instruct-hf & completion &  0.7766 & 0.884& 0.938 &0.507\\
    pfnet/plamo-100b & completion & 0.7469 & 0.861 & 0.920 & 0.460 \\
    CohereForAI/c4ai-command-r-plus & completion & 0.7365 & 0.818 & 	0.913 & 0.478 \\
    nvidia/nemotron-4-340b-instruct & completion & 0.7175 & 0.816 & 0.908 & 0.429 \\
    meta-llama/Meta-Llama-3.1-405B & completion & 0.6759 & 0.767 & 0.892 & 0.368\\
    google/gemini-1.5-pro-001 & chat &  0.6745& 0.666 & 0.980 & 0.377\\
    tokyotech-llm/Swallow-MX-8x7b-NVE-v0.1 & completion & 0.6368 & 0.753 & 0.870 & 0.287\\
    stabilityai/japanese-stablelm-base-beta-70b & completion & 0.6202 & 0.733 & 0.848 & 0.280 \\
    openai/gpt-35-turbo & chat & 0.6136 & 0.658 & 0.944 & 0.239 \\
    stockmark/stockmark-100b & completion & 0.5971 & 0.709 & 0.842 & 0.240\\
    meta-llama/Meta-Llama-3.1-70B & completion & 0.5659 & 0.665 & 0.822 & 0.211\\
    Qwen/Qwen-72B-Chat & completion & 0.5002 & 0.614 & 0.716 & 0.171\\
    mistralai/Mixtral-8x22B-v0.1 & completion & 0.4050 & 0.517	 & 0.615 & 0.084\\
    mistralai/Mixtral-8x7B-Instruct-v0.1 & completion & 0.3914 & 	0.488 & 	0.636 & 	0.050\\\hline
    \end{tabular}
    \label{tab:score-table}
\end{table*}

\end{document}